\pdfoutput=1

\documentclass[11pt]{article}

\usepackage[]{acl}

\usepackage{times}
\usepackage{graphicx}
\usepackage{caption}
\usepackage{subcaption}
\usepackage{booktabs}
\usepackage{latexsym}
\usepackage{tipa}
\usepackage{enumitem}
\usepackage{adjustbox}

\usepackage[T1]{fontenc}

\usepackage[utf8]{inputenc}

\usepackage{microtype}

\newcommand{\Figure}[1]{Fig.~\ref{#1}}

\newcommand{\Sectref}[1]{Section~\ref{#1}}

\title{\textit{Dim Wihl Gat Tun}: The Case for Linguistic Expertise in NLP for Under\-documented Languages}

\newcommand{\indp}{\textrm{\textipa{Z}}}
\newcommand{\ubc}{\textrm{\textipa{Q}}}

\author{Clarissa Forbes$^{\indp}$ ~~~
 Farhan Samir$^{\ubc}$ ~~~ Bruce Harold Oliver$^{\ubc}$ ~~~ Changbing Yang$^{\ubc}$ \\ {\bf Edith Coates}$^{\ubc}$ ~~~
 {\bf Garrett Nicolai}$^{\ubc}$ ~~~ {\bf Miikka Silfverberg}$^{\ubc}$ \\
$^{\indp}$Independent Researcher~~~ $^{\ubc}$University of British Columbia\\
 $^{\indp}$\texttt{forbesc@alumni.ubc.ca}~~~$^{\ubc}$\texttt{first.last@ubc.ca}}
  
\usepackage{linguex}

\newcommand{\Gitg}{\smash{\underline{g}}}
\newcommand{\Gitk}{\underline{k}}
\newcommand{\Gitx}{\underline{x}}

\begin{document}
\maketitle
\begin{abstract}
Recent progress in NLP is driven by pretrained models leveraging massive datasets and has predominantly benefited the world’s political and economic superpowers. Technologically underserved languages are left behind because they lack such resources. Hundreds of underserved languages, nevertheless, have available data sources in the form of interlinear glossed text (IGT) from language documentation efforts. IGT remains underutilized in NLP work, perhaps because its annotations are only semi-structured and often language-specific. With this paper, we make the case that IGT data can be leveraged successfully provided that
target language expertise is available. We specifically advocate for collaboration with documentary linguists. Our paper provides a roadmap for successful projects utilizing IGT data: (1) It is essential to define which NLP tasks can be accomplished with the given IGT data and how these will benefit the speech community. (2) Great care and target language expertise is required when converting the data into structured
formats commonly employed in NLP. (3) Task-specific and user-specific evaluation can help to ascertain that the tools which are created benefit the target language speech community. We illustrate each step through a case study on developing a morphological reinflection system for the Tsimchianic language Gitksan.     
\end{abstract}

\section{Introduction}


Progress\footnote{\textit{Dim wihl gat tun} - ``This is what the people should do''}\footnote{First two authors contributed equally.} in NLP research has primarily manifested in tools for the world's political and economic superpowers \citep{blasi2021systematic}, and it
is unclear how we can build more inclusive language technologies. 
Even multilingual pretraining methods \citep[e.g.,][]{liu2020multilingual,artetxe-etal-2018-unsupervised}, capable of producing effective models in the absence of large annotated training datasets require unannotated corpora that are prohibitively large for 90\% of the world's languages \citep{joshi2020state}. 

Nevertheless, many languages in this 90\% have a body of resources.
Language documentation and linguistic fieldwork are an ongoing task worldwide, and many resources continue to be developed in these traditions \citep{bird2020decolonising}.
We have access to wordlists, bilingual dictionaries for over 1000 languages \citep{Wu2020MultilingualDB}, aligned speech recordings for over 700 languages \citep{8683536}, multi-parallel texts for 1600+ languages \citep{mccarthy-etal-2020-johns}, and knowledge of related languages \citep{haspelmath2005world}.
Indeed, researchers have leveraged these resources to build impressive, useful computational systems for 
multilingual morphological analyzers \citep{nicolai-yarowsky-2019-learning}, adapting pretrained language models for over 1000 languages \citep{ebrahimi-kann-2021-adapt}, and building massively multilingual speech recognition systems \citep{adams-etal-2019-massively}, among others.

\begin{figure}
    \centering
    \adjustbox{width=\columnwidth}{
    \includegraphics{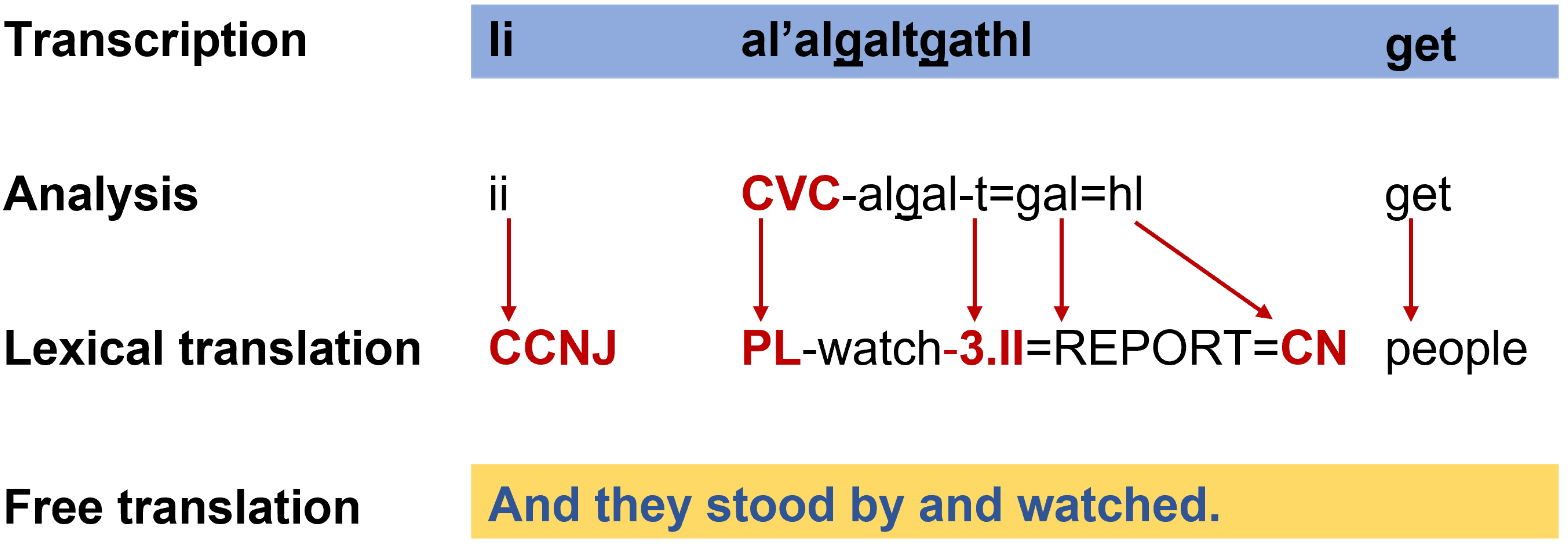}}
    
    \caption{An example of Gitksan interlinear glossed text (IGT). The text contains four levels of annotation: (1) An orthographic transcription, (2) A segmentation into normalized component morphemes (CVC refers to the reduplicated segment \textit{al'}), (3) an interlinear gloss and (4) an English translation.}
    \label{fig:IGT}
\end{figure}

There are additional language documentation resources which have yet to be fully leveraged in the aim to produce more inclusive language technology. Interlinear glossed texts (IGTs) depicted in Figure \ref{fig:IGT} are semi-structured texts which comprise not only monolingual corpus data (e.g. \textit{al'al\underline{g}alt\underline{g}athl}) but also morpheme-level segmentations (e.g. \textit{CVC{\textasciitilde}al\underline{g}al-t={\Gitg}at=hl}),  glosses for component-morphemes (e.g. \textit{PL{\textasciitilde}watch-3.II=REPORT=CN}), word alignment information \citep{zhao-etal-2020-automatic}, and free translations.
IGTs remain a major annotated datatype produced in the course of linguistic fieldwork: examples are
continuously digitized in large databases for hundreds of languages \citep{lewis2010developing}, and entire corpora of IGT are periodically published in volume series such as \textit{Texts in Indigenous Languages of the Americas}.
They have the potential to serve as training data for a wide variety of computational systems including bilingual lexicons, morphological analyzers, dependency parsers, part-of-speech taggers, and word-aligners \citep{georgi-thesis}.
Yet while they are accessible, they remain severely 
underutilized for these purposes. 

Part of the general hesitancy in adoption of IGT as training data may lie in the fact that the annotation format is only semi-structured and often language-specific.
While the general IGT format is governed by the Leipzig glossing rules \citep{comrie2015leipzig}, there remains significant flexibility for the annotator to customize tags and conventions for any given language.
This makes IGT challenging as a format for training supervised NLP models. 

With this paper, we make the case that IGT data can be leveraged in NLP research and language applications for speech communities, provided that target language expertise is available.
Specifically, we argue that it is essential to collaborate  with documentary linguists who are familiar with the language-specific annotations in the IGT data in order to leverage the data for NLP tasks. This may furthermore provide a foundation for co-designing language technologies with a given speech community \citep{bird2020decolonising}.

\begin{figure*}
    \centering
    \includegraphics[scale=0.5]{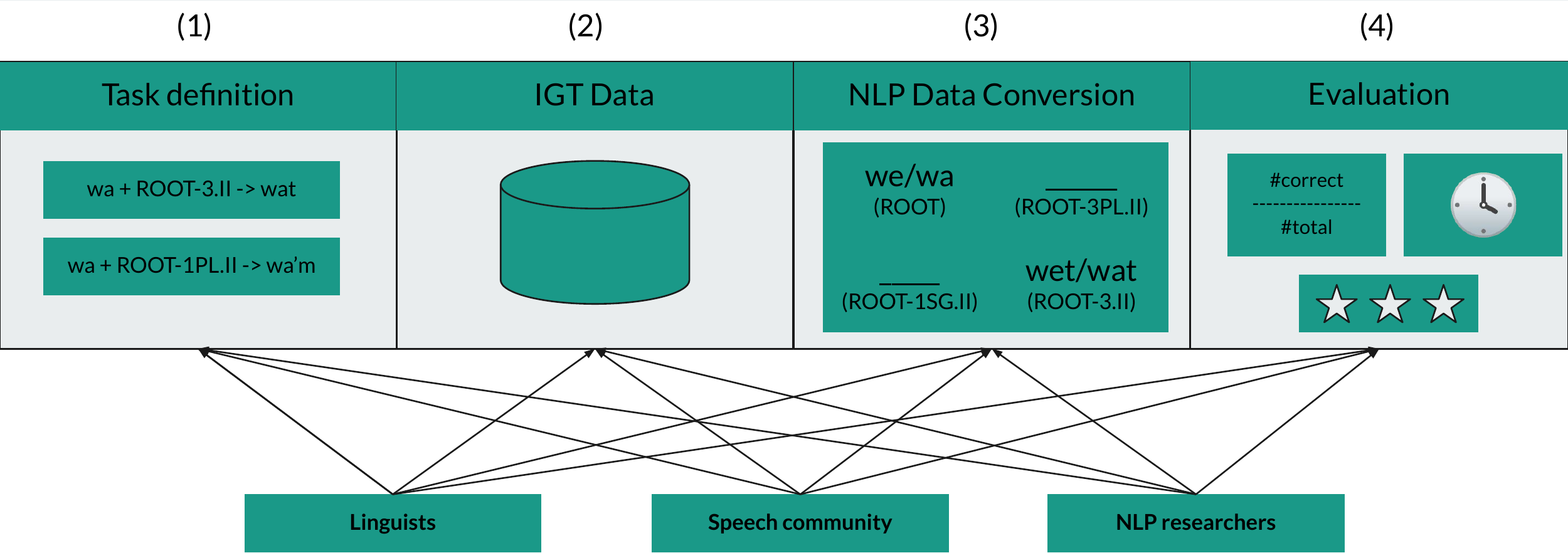}
    \caption{A roadmap for incorporating Interlinear Glossed Text (IGT) data for building more inclusive language technology. (1) We first need to define what NLP tasks can be accomplished with a given set of IGT data and whether they are valuable to the speech community (see \Sectref{sec:project-context}). (2) Next, we need to gather the relevant IGT data that was created during linguistic fieldwork with the speech community (see \Sectref{sec:project-context}). (3) Next, the IGT data needs to be converted to a structured format amenable for NLP formats. (4) The model needs to be evaluated not only in terms of standard NLP model selection metrics but also for efficacy for end-users such as efficiency in time-savings and usability (see \Sectref{sec:modeling}). Crucially, all three stakeholders -- speech community members, NLP researchers, and linguists -- should be involved throughout the process.}
    \label{fig:collabschematic}
\end{figure*}

Our paper provides a roadmap, portrayed in \Figure{fig:collabschematic}, for navigating three areas of significant uncertainty that arise when incorporating IGT data for inclusive language technology. First, we need to define what NLP tasks can be accomplished with a given set of IGT data, and whether they are of value to the speech community.
Second, after selecting useful tasks, we will need to preprocess the data, potentially by converting it to a structured format commonly employed in NLP tasks. 
Finally, we need task-specific and user-specific evaluation procedures in order to be explicit about the failure modes of the technology, as it is ultimately being developed for end users like speakers and linguists rather than solely comparison with other researchers.  

We focus on the first two of these areas, forwarding our argument through a case study on developing a morphological reinflection system for the Gitksan language (\Sectref{sec:project-context}) that has applications in language teaching.

\begin{figure}

\label{fig:igt-gitksan-example}
\end{figure}

\section{Background} \label{sec:background}

\subsection{NLP for Underdocumented Languages} \label{sec:nlp-under-documented} 

Computational work on under\-documented and low-resource languages has accelerated in recent years due to increasing recognition of both the role of NLP in language preservation as well as dedicated workshops like ComputEL \cite{computel-2021-use}, AmericasNLP \cite{americasnlp-2021-natural} and SIGTYP \cite{sigtyp-2021-typology}. Most of this work aims to assist in language documentation and revitalization, with machine translation being another important research area. \newcite{mager-etal-2018-challenges} and \newcite{littell2018indigenous} present surveys of existing NLP tools for the North American Indigenous languages, many of which are under\-documented, and discuss core challenges: morphological complexity, limited training data, and dialectal variation. 

Several authors have trained NLP models on IGT to accelerate language documentation, with automatic glossing being a prominent research direction. The first approaches simply memorized earlier glossing decisions and enabled the annotator to re-use these later \cite{baines2009fieldworks}. Later approaches have relied on structured models like CRFs \cite{mcmillan2020automating}, RNN encoder-decoders \cite{moeller2018automatic} and transformers \cite{zhao-etal-2020-automatic} to generate glosses for unseen tokens. NLP techniques can also be used to generate inflection tables from IGT \citep{moeller2020igt2p}. These find applications both in language documentation and language education, often to facilitate the production of more IGT data. A related approach is to generate morphological analyzers using IGT as a starting-point \cite{zamaraeva2016inferring,wax2014automated}. 

Several papers discuss challenges related to IGT as a data type. One of the principal concerns is the noisiness of the annotations \cite{moeller2020igt2p}. This problem is compounded by the fact that annotation schemas employed by linguists preparing IGT tend to be idiosyncratic\footnote{These systems are well motivated but unlikely to be easily comparable with other annotation schemas.} and often lack internal consistency \cite{baldridge2009well,palmer2009evaluating}. The design of annotation standards is important: \newcite{zhao-etal-2020-automatic} note that this can have an impact on the performance of glossing systems. \newcite{mcmillan2020automating} notes a further challenge: IGT often includes not only morphological information, but also syntactic, semantic, and pragmatic annotations, which can be much harder to learn in low-resource settings. 

In addition to challenges in the IGT data type itself, there are other challenges in NLP applications for under\-documented languages. \newcite{ward2003call} discuss many problems related to development of computer-assisted language learning for endangered languages: lack of orthographic standards, limited resources, and limited documentation of the language. \newcite{van2019future} also discuss NLP tools that can be helpful for documentation of low-resource languages, but they note that restrictive licenses can often be problematic for engineering.

\subsection{The Gitksan Language}
\label{sec:gitksan}
The Gitxsan are one of the Indigenous peoples of the northern interior region of British Columbia, Canada.
Their traditional territories consist of upwards of 50,000 square kilometers of land in the upriver Skeena River watershed area.
Their traditional language, called Gitksan in the linguistic literature, is the easternmost member of the Tsimshianic family, which spans the entirety of the Skeena and Nass River watersheds to the Pacific Coast.

Today, Gitksan is the most vital Tsimshianic language, but is still critically endangered with an estimated 300-850 speakers \citep{fpcc2018sotl}.
Community revitalization efforts are underway but are primarily undertaken by individuals on an ad-hoc basis.
Initiatives include regular in-school language programming, a few adult language courses, a successful language immersion camp, and several Master-Apprentice pairs.

Linguistic documentation on Gitksan and the Tsimshianic languages has been going on intermittently since the 1970s, including the drafting of a never-published grammar \citep{rigsby1986} and waves of formal phonological, syntactic, and semantic work over the past thirty years.
There are several community-developed wordlists and workbooks, but no comprehensive dictionary, grammar, or pedagogical curriculum.
There is an accepted orthography \citep{hindlerigsby}, and a talking dictionary mobile app in active use by the community (\textit{Mother Tongues Dictionaries}, formerly \textit{Waldayu}; \newcite{littell2017waldayu}).


Other computational studies interact with the active documentation efforts surrounding Gitksan to produce new frameworks and resources. 
\newcite{dunham-etal-2014} present a database structure for hosting audio and transcribed data in language documentation contexts, adopted for Gitksan and eight other under\-documented languages.
\newcite{littell2017waldayu} present a dictionary interface which is capable of fuzzy search. They mention this specifically as a way to increase accessibility in a setting where orthographies have not been standardized or where many users are language learners. \newcite{forbes-etal-2021-fst} present a finite-state morphological analyzer for Gitksan; they test coverage across different dialects of Gitksan and use handcrafted rules to increase coverage for spelling variants. 

\subsection{Constructing a Gitksan Pedagogical Application from IGT Data}
\label{sec:project-context}

Our project generates language learning exercises for Gitksan grammar. The need for these exercises was identified in discussions with documentary linguists working on Gitksan (the task definition step in Figure \ref{fig:collabschematic}). 
Specifically, our goal is to automatically generate exercises for noun and verb inflection. 
As source material, we use Gitksan IGT data collected by linguists at the University of British Columbia for language documentation purposes (the data step in Figure \ref{fig:collabschematic}). Examples of this data are shown in Figure \ref{fig:IGT} and Appendix \ref{app:sampleIGT}. 

Due to extensive morphological annotation, IGT provides a valuable starting point for our work. However, the annotations are far too detailed for our purposes---many derivational affixes are annotated in the data (further discussed in Section \ref{sec:granularity}). These are irrelevant and can be downright harmful for grammar exercises. 
To remedy this misalignment between the raw IGT data and our NLP task, we collaborate with Gitksan documentary linguists to identify a set of inflected forms with clearly defined grammatical function, while discarding derivational morphology. We then convert the IGT data into a set of inflectional paradigms (the data conversion step in Figure \ref{fig:collabschematic}). We further discuss this conversion process in Sections \ref{sec:standards} and \ref{sec:variation}. Since the inflectional paradigms sourced from corpora are sparse,\footnote{Due to the Zipfian distribution of language \cite{blevins2017zipfian}.} we train models to fill in missing forms (\Sectref{sec:modeling}). This is more widely know as the Paradigm Cell-Filling Problem (PCFP)  \citep[e.g.,][]{Silfverberg2018AnEA}. 
We then evaluate the system on it’s capacity to automatically generate inflections, and discuss limitations of our current evaluation procedure (the evaluation step in Figure 2).


\section{Challenges in Incorporating IGT into NLP Research}
\label{sec:preprocessing}

Because tokens in IGT are already segmented and annotated, it forms an ostensibly convenient starting-point for further processing and token-based grouping. In many ways, IGT is, however, a challenging data type for use in pedagogical and NLP applications.
This section presents three specific challenges posed by IGT data when NLP techniques are applied.
First, while IGT will contain a wealth of useful information for NLP models, it might also contain information which is far too fine-grained for automatic learning purposes, at least given the quantity of data which are available.
Second, IGT often contain idiosyncratic or language-specific conventions which may not be easily converted to or represented in standardized frameworks.
Third, because IGT is used as a device for language documentation, it will often contain dialectal variation, an important meta-characteristic which in aggregate cannot be easily distinguished from other types of variation or spelling errors.
We argue that handling these issues for successful data preprocessing requires consultation with linguistic experts, and exemplify with instances from the Gitksan IGT and our use-case.




\subsection{Annotation Granularity}
\label{sec:granularity}


Documentary linguists' goals when annotating IGT is to present an accurate representation of the surface phonology and morphology of a given utterance, as well as the syntactic and semantic information contributed by its component morphemes, with fine attention to detail given the rarity and value of the data. 
This goal of providing fine-grained annotations and transcriptions, however, can be in conflict with the NLP research aim of building models that can generalize in the real world (i.e., future elicited linguistic data). The fine-grained details are often extraneous for the purposes of building NLP models, and can counterproductively act as noise that makes learning systematic patterns more difficult.

As an example of this mismatch in disciplinary goals, consider the sample IGT token in \Next. 

\ex.\glll  maaxwsxwa \\
            maaxws-xw-a \\
            fallen.snow-\textsc{val}-\textsc{attr} \\
            `white'

In this token, the productive stem is deconstructed into a historical root (\textit{maaxws})
and a derivational suffix (\textit{-xw})---along with an inflectional affix (\textit{-a}).
It is unclear from the input that the most readily recognizable lexical stem in this form is the larger unit \textit{maaxwsxw} `white, snow-colored', and that the internal boundaries within that stem reference etymological and derivational information not relevant to the typical NLP task.
The derivational and inflectional affixes are not differentiated in IGT.\footnote{For an English analogue, consider splitting the lexicalized verb \textit{enforce} into a prefix \textit{en-} and root \textit{force}. The \textit{en-} prefix is recognizable, but not productive or relevant to inflection tasks.} 

At first glance, it might seem reasonable to train an NLP model to automatically generate such a gloss for Gitksan input words in an effort to accelerate language documentation. 
While this remains one of the most common NLP tasks associated with IGT, it may be difficult for models to deliver high performance if the IGT input, like Gitksan's, contains a substantial proportion of derivational and etymological information, since this information is lexical and unpredictable.

Collaboration with documentary linguists, in addition to being important when a project aims to improve the documentary linguistic workflow, can be useful for identifying these aspects of the data which may be less valuable to learn. This information can be applied in data preprocessing to improve model performance given data scarcity.
For the token in \Last, an alternative segmentation \textit{maaxwsxw-a} into a word stem and a productive inflectional affix \textit{white-ATTR} is more amenable to both automated labeling and inflection tasks, particularly in low-resource conditions. 
Furthermore, reference to derivational information is unnecessary in our use case of performing automated inflection for use in a pedagogical application.
We collaborated with documentary linguists familiar with Gitksan to manually filter morphology into derivational versus inflectional, to determine whether an affix should be classed as part of a lexical stem or should signal a paradigm cell in the inflectional template.
This allowed derivational morphology to be effectively excluded before we moved to the paradigm cell-filling task.
This filtering process was non-trivial, requiring solid understanding of the target language, its description, and its vocabulary.

\subsection{Using Existing Annotation Standards}
\label{sec:standards}


The annotation schemas employed in IGT are often idiosyncratic \cite{palmer2009evaluating,comrie2015leipzig}, which typically makes them better suited for language documentation than  NLP tasks.
When aiming to leverage IGT data for use in NLP tasks, we must then consider on a case-by-case basis whether it is more beneficial to convert the IGT data to an NLP-standard format, or work with the IGT annotations largely as-is, adapting them to our specific needs.
Relevant to this decision are factors such as how labor-intensive the conversion will be, how well the standard format accommodates linguistic information that has been detailed in the IGT, and whether conversion of the dataset to the standard format aligns with specific project goals and speech community interests.

The possible format that we consider for annotating inflection tables is the Unimorph standard \cite{mccarthy-2020-unimorph,sylak2016composition}, a popular schema for annotation of inflectional morphology that can facilitate cross-lingual transfer by enabling language-independent annotations.
Ultimately, we opted to adapt the Gitksan IGT to our specific needs after determining that conversion would be extremely labor-intensive, and that several types of information in the Gitksan IGT could not be represented in the UniMorph standard.
We present three of the most significant issues:



\paragraph{1. Part-of-Speech} The Unimorph standard relies on part-of-speech (POS) tags as a major component of word form annotation. However, POS information is frequently not annotated in IGT \cite{moeller2020igt2p}, and no POS information was included in our Gitksan IGT.

For some underdocumented languages, POS information requires substantial experience and manual attention to annotate. For example, our target language Gitksan displays considerable category flexibility, meaning that syntactic and morphological behavior can cross word class boundaries. 
In Gitksan, the inflectional paradigms of nouns and verbs overlap substantially. As an example, agreement markers can affix to both nouns and verbs, conveying a number of functions. Some are exemplified in \Next.
As a consequence, in Gitksan it is difficult to use morphological inflection to deduce a lexeme's POS.

\ex. Forms with \textit{-'y} (\textsc{1sg} series II)
\a. \textit{hlguuhlxwi\textbf{'y}} - my child (\textsc{possr})
\b. \textit{yee\textbf{'y}} - I walked (\textsc{abs})
\b. \textit{t'agi\textbf{'y}} - x forgot me (\textsc{abs}, dependent)
\b. \textit{t'agi\textbf{'y}} - I forgot x (\textsc{erg})

In addition, Gitksan nouns and verbs are syntactically flexible, meaning that Gitksan nouns can function as verbs in text, and vice versa. For example, a noun \textit{\Gitg anaa'w} `frog' can be used predicatively without a copula in main verb position in the sentence \textit{Hlaa ap \textbf{\Gitg anaa'w}i'y} `I'm a frog now'. It takes absolutive inflection when it does so.
Due to this morphological and syntactic flexibility, a 1SG-inflected noun like \textit{\Gitg anaa'wi'y} could be annotated two ways in UniMorph depending on the context (\texttt{frog;PSS1S} versus \texttt{frog;1SG;ABS}\footnote{Other clause type features would be required here but it remains unclear how best to represent Gitksan's clause-typing system with UniMorph labels.})---yet in the IGT, they are uniformly annotated as \texttt{frog-1SG.II}.
Reviewing the contextual function of every noun and verb in the IGT dataset to apply the appropriate UniMorph tags would require an infeasible amount of expert reannotation. 

\paragraph{2. Inflection vs. derivation} 
Unimorph postulates a strict division into inflectional and derivational morphology (and only annotates inflectional morphology). The IGT format has no such division, because it can be used to represent morphology at any level of granularity the annotator wishes.

We have mentioned in \Sectref{sec:granularity} that determining the difference between inflectional and derivational morphology from IGT input is non-trivial. 
For example, the Gitksan morpheme \textit{-xw} has a variety of uses which might be considered more derivation-like (D) or more inflection-like (I).

\begin{itemize}[noitemsep]
    \item Creating intransitive predicates from nouns: \textit{os\textbf{xw}}~`have~a~dog' from \textit{os}~`dog' (D)
    \item Marking inchoatives: \textit{mit\textbf{xw}}~`be~full' vs.\ causative \textit{midin}~`fill' (D)
    \item Marking passives: \textit{jap\textbf{xw}}~`be~made' from transitive \textit{jap}~`do,~make' (D?)
    \item Marking verbs with certain preverbs: \textit{sik'ihl~huut\textbf{xw}}~`try~to run~away' vs.\ \textit{huut}~`run~away' (I?)
    \item Optional in some possessives: \textit{la\Gitx yip\textbf{xw}si'm} `your.pl land' vs.\ \textit{la\Gitx yipsi'm} `your.pl land' (?)
\end{itemize}

This morpheme's uses and degree of productivity are still little-understood, so its status as inflectional or derivational remains unclear.\footnote{Elsewhere, some linguistic descriptions present cases of morphology which do not fit into conventional delineations of the inflectional/derivational divide, such as plural/pluractional markers in Halkomelem Salish \citep{wiltschko2008}.} 
For now, we provisionally exclude this morpheme from our inflection tables as `derivational'. In a UniMorph system, this morpheme's exclusion or inclusion in the annotation would constitute a prematurely strong claim about whether it was inflectional, and the tagset used to annotate it likewise a prematurely strong claim about its function.




\paragraph{3. Clitics} 
Gitksan is rich in clitics, annotated with the equals sign in IGT `='. Their attachment is determined by prosodic and linear factors. 
Pre-nominal clitics are illustrated in example \Next.
\ex.\glll   Giigwis Maryhl  \Gitg ayt. \\ 
            giikw-i[-t]=s Mary=hl \Gitg ayt \\
            buy-\textsc{tr}-\textsc{3.ii}=\textsc{pn} Mary=\textsc{cn} hat \\
            `Mary bought a hat.'

In the example above, the proper noun clitic \textit{=s} attaches to the verb but is syntactically associated with \textit{Mary}. The common noun clitic \textit{=hl} attaches to \textit{Mary} but is associated with \textit{\Gitg ayt} `hat'. Since UniMorph does not annotate such cross-token dependencies (or other clitics), this central feature of Gitksan cannot be represented.


\paragraph{Recommendations} Current computational morphology research relies heavily on standardized tagsets like UniMorph, in particular for crosslingual transfer \citep{anastasopoulos-neubig-2019-pushing}. However, these formats can be either labor-intensive or impossible to apply to under\-documented language datasets, depending on the idiosyncratic conventions of a given IGT and language-specific factors. Our understanding of the language may not be sufficiently mature to implement some of UniMorph's strict requirements, or important phenomena may fall outside of the defined scope of UniMorph. We recommend that NLP projects on under\-documented languages collaborate with language experts to determine where language-agnostic data formats can be applied, and to design project-specific data formats as needed.

\subsection{Dialectal variation}



\label{sec:variation}
Dialectal variation is a pervasive feature of languages worldwide, from English \citep[consider African-American English and Standard American English;][]{blodgett-etal-2016-demographic} to Arabic \citep[consider Modern Standard Arabic and the Doha dialect;][]{Kumar2021MachineTI}. Many Indigenous languages of North America also exhibit vast dialectal variety, with significant variance in the level of mutual intelligibility between languages and dialects \citep[Ch.6]{mithun2001languages}. 

Although Gitksan has an estimated fewer than 1K speakers, each village has a different way of speaking, and the speech community recognizes two salient dialects (Eastern/Upriver and Western/Downriver). Gitksan dialectal variation is typically reflected in written materials due to the lack of a widely-adopted orthographic standard which would `flatten' it.\footnote{Linguistic description frequently aims to record dialectal and even speaker-level variation. Our datasets are based on IGT data which explicitly annotates such variation in the orthographic representation.} For many under\-documented languages, written orthographies have been in use for a relatively short period of time, and communities place different levels of emphasis on literacy and standardization versus conversational fluency. As a consequence, orthographic conventions can vary widely across dialects and writers in low-resource and underdocumented language contexts.

It is desirable in building inclusive language technology to accommodate and reflect variation, rather than aim to model a homogenous standard form of the language. In building pedagogical resources for language revitalization, we furthermore need to mindfully consider potential data biases as well as what kinds of variation are presented to the user, to avoid implictly suggesting that certain dialects favored for preservation and teaching, which risks reinforcing or creating negative social hierarchies \citep{Demszky2021LearningTR}. 

The first step to ensuring dialectal fairness and appropriate handling of variation in NLP applications is to understand what types of variation are at play, and in particular what dialect a given token belongs to.
This allows us to proactively control what data is presented to a user and, for example, ensure that data from different dialects is not mixed together inappropriately.
This task is non-trivial: expertise in the language is crucial in order to determine what types of variation are dialectal, and which are idiosyncratic or purely orthographic, including typos and spelling errors. As an example from Gitksan, \textit{gat} and \textit{get} are highly salient East/West dialect variants, while \textit{hun} and \textit{hon} are less-salient variants within the Eastern dialect; \textit{amxsiwaa} and \textit{amxsiiwaa} are two non-dialectal variants of the same word (spelling error/variant), while \textit{sipxw} and \textit{siipxw} are different lexemes.\footnote{In IGT the gloss cannot always be used to differentiate lexemes. Depending on the convention, the same lexeme may appear with different glosses in different contexts (e.g.\ \textit{'wa}: `find' or `reach'), and different lexemes may have the same gloss (e.g.\ \textit{yoo{\Gitk}} and \textit{gup}: `eat', which differ on other grounds -- transitivity). The latter forms which share a gloss must also be differentiated as lexical variants, not dialectal variants.}
Presently, we include all lexeme variants as separate entries in our inflection tables, enabling us to represent all dialects during training.

\paragraph{Recommendations} Distinguishing between different types of variation in the source material is a challenging task but also a crucial one. Expertise in the target language and dialects is required for classifying types of variation, and so language experts are a vital asset for this process. Documentary linguists or community members may have direct information about the dialectal background of speakers that are represented in the data, which is useful for modeling, and will likely have information about how dialectal variation is viewed in the speech community (e.g.\ it may be highly politicized), which is important for application design.


Variation is not only an important issue when constructing datasets. It is also essential to evaluate the final model's performance according to the principle of dialectal fairness \citep{Choudhury2021HowLF}
Recently, measures for dialect fairness have emerged in the NLP community: 
 \newcite{faisal2021sd} and \newcite{Kumar2021MachineTI} advocate for computing performance separately for each dialect rather than computing a single macro average performance figure over distinct dialects. They also propose to use standard deviation between system performance on different dialects and the generalized entropy index \cite{speicher2018unified} as measures for dialectal unfairness which we naturally want to minimize.

\section{Steps toward Building a Language Learning Application}
\label{sec:modeling}
The inflectional paradigms collected from the adapted IGT corpus are overly sparse for 
automatically generating pedagogical exercises. To automatically fill in these
paradigms, an example of which is shown in Appendix B, we train and
evaluate a morphological reinflection system.\footnote{Code and data for this experiment is available at \url{https://github.com/smfsamir/gitksan-data}.}

\paragraph{Data}
We train and test reinflection models on the Gitksan morphological paradigms described in \Sectref{sec:preprocessing}. 
We generate three splits of the data from our complete set of paradigms: train ($N=858$ word forms), validation ($N=302$ word forms), and test ($N=124$ word forms) data splits.

\paragraph{Training} We form training pairs by using the given forms in each table and learn to reinflect each given form
in a table to another given form in the same table, following \citet{Silfverberg2018AnEA}. Model parameters are shown in Appendix \ref{appendix:hyp}.
\paragraph{Evaluation} During test time, we predict forms for missing slots based on each of the given forms in the table and take a majority vote of the predictions. We evaluate accuracy on the test set by counting the number of the $124$ forms that were correctly predicted. We find that the Transformer model generates   87.09\% of the test forms correctly.

\paragraph{Analysis.} 
Our model provides strong performance when measured by the standard metric of accuracy, in particular considering that it is trained on only $858$ examples. Accuracy, however, only provides one perspective on the efficacy of the model \citep{ethayarajh-jurafsky-2020-utility}.
The appropriate evaluation of the system is highly context dependent: For our goal of generating language learning exercises, we want to evaluate whether our system and automatically generated grammar exercises allow for more effective language learning; raw accuracy gleans little insight to the effectiveness of the system for this goal.  
If in contrast our goal was to facilitate language documentation, 
we would want to evaluate whether the model gives an overall significant reduction in documentation effort
---
this largely depends on whether the automatic annotations are of sufficient quality that correcting remaining errors takes less time than annotating all the data from scratch. Further research, in collaboration with documentary linguists and the speech community, is required to determine whether our system can achieve the desired goals of building more practical, inclusive language technology.

\section{Discussion}

\paragraph{Incorporating IGT data for NLP} Language documentation provides a valuable data source for many so called ``left-behind'' languages \citep{joshi2020state}, which lack traditional annotated and unannotated NLP datasets. For example, IGT data can be used to train systems for morphological inflection, segmentation and automatic glossing, among other applications. Nevertheless, the annotations in IGT are rarely ideally suited for typical NLP tasks, and may need to be significantly adapted. This will typically be hard without extensive knowledge of the target language and annotation conventions which were employed when the IGT data were generated. 
Linguists and community language experts are well-positioned to address questions related to IGT usability, the structure of the target language, variation in the data, and other annotations in the source data. Collaboration with language experts is not only vital for successful data preprocessing and conversion to the formats required for the typical NLP task, but can also naturally help define research goals and drive the project toward them.

\noindent \textbf{Inclusive Research Goals} NLP technologies for under\-documented languages have the capacity to speed up language documentation \citep[e.g.,][]{anastasopoulos2019computational}; assisting language revitalization \citep[e.g.,][]{Rijhwani2020OCRPF,lane2020bootstrapping}; and creating digital infrastructure \citep[e.g.,][]{anastasopoulos-neubig-2019-pushing}.
These high-level goals are only a part of what it may mean to create inclusive language technology.
Equally valuable as a research goal may be \textbf{inclusion}: for speech communities to be acknowledged and engaged in the course of the the research project.\footnote{Under\-documented languages are often the cultural heritage of typically marginalized peoples, sometimes with a history of their data being exploited for political or commercial purposes. NLP research without community involvement may feel like a continuation of this pattern.}
We encourage NLP projects on low-resource, minoritized, and/or endangered languages to begin by understanding the speech community context, proceed with community collaboration or endorsement, and ultimately produce concrete benefits that speech communities recognize. This might include outcomes for language teaching and pedagogy, or training opportunities in technology or research.

Evaluation methods can be compiled which address NLP researchers, linguists, and communities' overlapping and divergent goals.
For example, pedagogical tools can be directly evaluated for dialect fairness and user/learner improvement.

\noindent\textbf{Practical Collaboration}
We suggest seeking out opportunities to collaborate directly with community members, in order to solicit their specific expertise when setting the research agenda (i.e.\ task definition) and conducting evaluation \citep{czaykowska2009research,bird2020decolonising}.
When the NLP researcher has no existing contact or history with the speech community, this can be pursued via collaboration with a documentary linguist with established community relationships and a similar desire to engage in this research model.
Recognize that in any collaboration, different individuals contribute different skills and experience (e.g.\ pedagogy, annotation, knowledge of community attitudes) and may have different goals and preferred ways of participating, which should simply be discussed within the partnership to ensure things run smoothly. 

\noindent \textbf{Research accessibility} In discussing inclusive language technologies, we also consider the accessibility of NLP workshops to speech communities, in particular where venues have a dedicated focus on low-resource languages. We note that such venues are often inaccessible to communities due to factors such as the cost of registration. Similarly-oriented workshops in linguistics (e.g.~SAIL, WSCLA, family-specific conferences) typically have a tiered registration structure enabling community members to attend for free or minimal cost (e.g.\ \$25). It is worth recognizing that community members are research stakeholders, and ensuring that venues are open to their participation.

\section{Conclusion}

Although a majority of the world's languages lack the kind of large annotated and massive unannotated datasets which are used to train modern NLP models for high-resourced languages like English \citep{joshi2020state,blasi2021systematic}, many languages have other potential data sources such as language documentation data, which so far have remained under-explored. However, care must be taken when applying this type of data, which originally is not intended for NLP use. This is important to ensure that the resulting technologies actually achieve their intended goals like accelerated language documentation or genuinely helpful computer-assisted language learning.

Collaboration with linguists can provide the expertise necessary to engage in modeling with IGT data for underdocumented languages. Linguists can help define an NLP task with good value propositions, given their familiarity and connections with the speech community. They can provide guidance on navigating the IGT format so that we can extract the most useful information for the task at hand. Finally, they can assist in evaluating whether the model achieves appropriate performance on the speech community use cases, and provide feedback on metrics for model success and fairness across dialects.
Throughout the development process, documentary linguists and speech community members should be consulted. This will further a greater understanding of the source data and lead to more equitable and effective technologies.  

\section{Acknowledgements}
We want to thank Henry Davis, Lisa Matthewson and the Gitksan research lab at the Department of Linguistics at UBC for generous help with this project and access to Gitksan IGT data. We also want to thank for anonymous reviewers for valuable comments. We also want to thank Samantha Quinto for assisting with visual design. This research was supported by funding from the National Endowment for the Humanities (Documenting Endangered Languages Fellowship) and the Social Sciences and Humanities Research Council of Canada (Grant 430-2020-00793).
Any views/findings/conclusions expressed in this
publication do not necessarily reflect those of the
NEH, NSF or SSHRC.

\bibliography{anthology,custom}
\bibliographystyle{acl_natbib}

\appendix
\clearpage
\onecolumn
\section{Sample IGT data}
\label{app:sampleIGT}

The first four lines of a sample text from the Gitksan interlinear glossed text corpus. This example is revised from initial publication in \newcite{forbesetal2017}.

{

\ex.[]\glll
Dim mehldi'y wila wilhl win hii hagun bekwhl mismaaxwsxum get {\Gitg}o'ohl ts'ebim Gitwinhlguu'l gik'uuhl. \\
dim mehl-T-i-'y wila wil=hl win hii hogun bekw=hl CVC{\textasciitilde}maaxws-xw-m get {\Gitg}o'o=hl ts'ep-m Gitwinhlguu'l gi-k'uuhl \\
PROSP tell-T-TR-1SG.II MANR be/do=CN COMP initially toward arrive.PL[-3.II]=CN PL{\textasciitilde}fallen.snow-VAL-ATTR people LOC[-3.II]=CN community-ATTR Gitwinhlguu'l prior-year \\
I will tell about when the white men first came to Kitwancool long ago.

\ex.[]\glll
Ha'on dii 'nekw hlidaa bekwhl get dipun, ii sa{\Gitg}ayt{\Gitg}oodindiithl hli gedihl Gitwinhlguu'l.\\
ha'on dii 'nekw hli=da bekw=hl get dip=un ii sa{\Gitg}ayt-{\Gitg}ooda-in-diit=hl hli get-T=hl Gitwinhlguu'l\\
not.yet FOC long PART=SPT arrive.PL[-3.II]=CN people ASSOC=DEM.PROX CCNJ together-all.gone-CAUS2-3PL.II=CN PART people-T[-3.II]=CN Gitwinhlguu'l\\
Not long after these people arrived, they gathered together the people of Kitwancool.

\ex.[]\glll
Hasa{\Gitk}diit dimt mehldiit win hlaa dim sii ha'niijo{\Gitk}t {\Gitg}o'ohl win t'aahl {\Gitg}alts'ephl Gitwinhlguu'l.\\
hasa{\Gitk}-diit dim=t mehl-T-diit win hlaa dim sii ha-'nii-jo{\Gitk}-t {\Gitg}o'o=hl win t'aa=hl {\Gitg}al-ts'ep=hl Gitwinhlguu'l\\
desire-3PL.II PROSP=3.I tell-T-3PL.II COMP INCEP PROSP new INS-on-dwell-3.II LOC[-3.II]=CN COMP sit[-3.II]=CN container-community[-3.II]=CN Gitwinhlguu'l\\
They wanted to tell about the new place where the village of Kitwancool is to be.

\ex.[]\glll
'Nit sa{\Gitg}ootxwhl "government" siwatdiit, ii dim 'nii wenhl dim jo{\Gitk}hl aluugiget {\Gitg}o'ohl la{\Gitx} "reserve" siwatdiit.\\
'nit si-{\Gitg}oot-xw=hl *government si-wa-T-diit ii dim 'nii wen=hl dim jo{\Gitk}=hl aluu-CV{\textasciitilde}get {\Gitg}o'o=hl la{\Gitx} *reserve si-wa-T-diit\\
3.III CAUS1-heart-VAL[-3.II]=CN *government CAUS1-name-T[-TR]-3PL.II CCNJ PROSP on sit.PL[-3.II]=CN PROSP dwell[-3.II]=CN clearly-PL{\textasciitilde}people LOC[-3.II]=CN on *reserve CAUS1-name-T[-TR]-3PL.II\\
The plan of the so-called government was that they will have Indian people live on a so-called reserve.

}

\clearpage
\section{Sample inflection table}
\label{app:infl-table}
A Gitksan inflection table for {\it 'wa} (`to find, reach') generated from IGT and displayed in TSV format. Many cells in the table are empty since they were unattested in the IGT data.

\begin{verbatim}
ROOT	find	'wa	'wa	'wa
ROOT-SX	_	_	_	_
ROOT-PL	_	_	_	_
ROOT-3PL	_	_	_	_
ROOT-ATTR	_	_	_	_
ROOT-3.II	find-3.II	'wa-t	'wat	'wa-3.II
ROOT-PL-SX	_	_	_	_
ROOT-1SG.II	_	_	_	_
ROOT-2SG.II	_	_	_	_
ROOT-2PL.II	_	_	_	_
ROOT-3PL.II	find-3PL.II	'wa-diit	'wadiit	'wa-3PL.II
ROOT-1PL.II	_	_	_	_
ROOT-PL-3PL	_	_	_	_
ROOT-TR-3.II	find-TR-3.II	'wa-i-t	'wayit	'wa-TR-3.II
ROOT-PL-3.II	_	_	_	_
ROOT-PL-ATTR	_	_	_	_
ROOT-PL-2SG.II	_	_	_	_
ROOT-TR-1SG.II	_	_	_	_
ROOT-PL-3PL.II	_	_	_	_
ROOT-PL-1SG.II	_	_	_	_
ROOT-TR-1PL.II	find-TR-1PL.II	'wa-i-'m	'wayi'm	'wa-TR-1PL.II
ROOT-PL-1PL.II	_	_	_	_
ROOT-TR-2PL.II	_	_	_	_
ROOT-TR-3PL.II	_	_	_	_
ROOT-TR-2SG.II	_	_	_	_
ROOT-PL-TR-3.II	_	_	_	_
ROOT-PL-TR-2SG.II	_	_	_	_
ROOT-PL-TR-3PL.II	_	_	_	_
ROOT-PL-TR-1SG.II	_	_	_	_
ROOT-PL-TR-1PL.II	_	_	_	_
ROOT-PL-TR-2PL.II	_	_	_	_
\end{verbatim}

\twocolumn
\section{Fairseq parameters}
\label{appendix:hyp}
\paragraph{Model}We use the Fairseq \cite{ott2019fairseq} model implementation of Transformer \cite{vaswani2017attention}.  
Both the encoder and decoder have $4$ layers with 4 attention heads, an embedding size of $256$ and hidden layer size of
$512$. We train with the Adam optimizer starting of the learning rate at $0.001$. We chose the batch size ($400$) and maximum updates ($20000$) based on the highest accuracy on the development data.
\end{document}